\documentclass{article}
\usepackage{iclr2025_conference,times}

\usepackage{amsmath,amsfonts,bm}

\def\eqref#1{equation~\ref{#1}}

\def\1{\bm{1}}

\DeclareMathAlphabet{\mathsfit}{\encodingdefault}{\sfdefault}{m}{sl}
\SetMathAlphabet{\mathsfit}{bold}{\encodingdefault}{\sfdefault}{bx}{n}

\usepackage{hyperref}
\usepackage{url}
\usepackage{graphicx}
\usepackage{booktabs}
\usepackage{amsmath}
\usepackage{amssymb}
\usepackage{multirow}
\usepackage{array}
\usepackage{subcaption}
\usepackage{xcolor}
\iclrfinalcopy

\title{Structure-Token Evidence-Anchored Reasoning for Scientific Chart Understanding}

\author{Alberlucia Rafael Soarez, Camila Ferreira, Daniel Kim, Mariana Costa, Alejandro Torres \\
University of Brasilia \\
\texttt{alejandro.torres@iscon.edu.br}
}

\newcommand{\ours}{STEER}

\begin{document}
\maketitle

\begin{abstract}
Scientific charts encode quantities in axes, legends, and geometric marks, yet large vision--language models still treat them as natural photographs. Visual in-context examples do not expose the coordinate frame; unconstrained chain-of-thought can name a plausible number that was never read from a bar. We present \ours{} (Structure-Token Evidence-anchored Reasoning), which freezes a Llama-3.2-Vision encoder and inserts three modules: a chart structure graph encoder (CSGE) that binds ticks, legend items, and marks; evidence-anchored step reasoning (EASR) that forces every arithmetic step to cite a graph node; and weak-parser strong-reasoner alignment (WPSR) that uses a specialized table extractor only as a teacher of node attributes. On ChartQA, \ours{} reaches 82.70 average relaxed accuracy versus 80.16 for ChartGemma and 76.40 for a LLaVA-CoT backbone trained on the same mix. Gains widen on CharXiv reasoning (33.60 vs.\ 29.20 InternVL Chat V1.5) and ChartQAPro CoT (40.70 vs.\ 37.17 Qwen2-VL-7B), where OCR shortcuts disappear. Ablations show that dropping node serialization or numeric candidate constraints undoes most of the reasoning lift.
\end{abstract}

\section{Introduction}

Charts are the default carrier of quantitative claims in papers, filings, and policy briefs. A model that can answer ``which series peaked first?'' or ``what is the 2019--2021 ratio?'' would unlock scientific search and financial monitoring, including settings where numerical claims sit beside noisy social-media timestamps~\citep{hu2026ai}. Progress on ChartQA looks strong: specialist models report average relaxed accuracy above 80~\citep{masry2025chartgemma,DBLP:conf/acl/MasryLTJH22}. The same systems collapse once the chart is an arXiv figure~\citep{wang2024charxiv}, an unannotated plot~\citep{xu2023chartbench}, or a dashboard from a previously unseen publisher~\citep{masry2025chartqapro}. The failure is not generic visual question answering. It is a mismatch between how charts are drawn and how current large vision--language models (LVLMs) are trained.

Three practices dominate. First, visual in-context learning concatenates whole-image exemplars and hopes the query chart will be read by analogy~\citep{zhou2024visual}. Similarity in layout is a weak proxy for a shared coordinate frame: two bar charts can look alike while their $y$-axes differ by an order of magnitude. Second, free-form chain-of-thought~\citep{wei2022cot,DBLP:conf/iccv/XuJWLSSY25} writes a fluent rationale that need not point at a mark. ChartR shows that early extraction errors propagate through later arithmetic~\citep{chen2026chartr}; ChartAgent recovers some of that gap only by calling external drawing tools~\citep{kaur2026chartagent}. Third, chart-specialist parsers~\citep{liu-etal-2023-matcha,masry-etal-2023-unichart} extract tables well inside ChartQA's distribution and fail on ChartQAPro, where ChartGemma's CoT overall is 9.80~\citep{masry2025chartqapro}. A strong general LVLM and a weak specialist currently have no shared channel for node-level attributes.

\begin{figure}[tp]
\centering
\includegraphics[width=\linewidth]{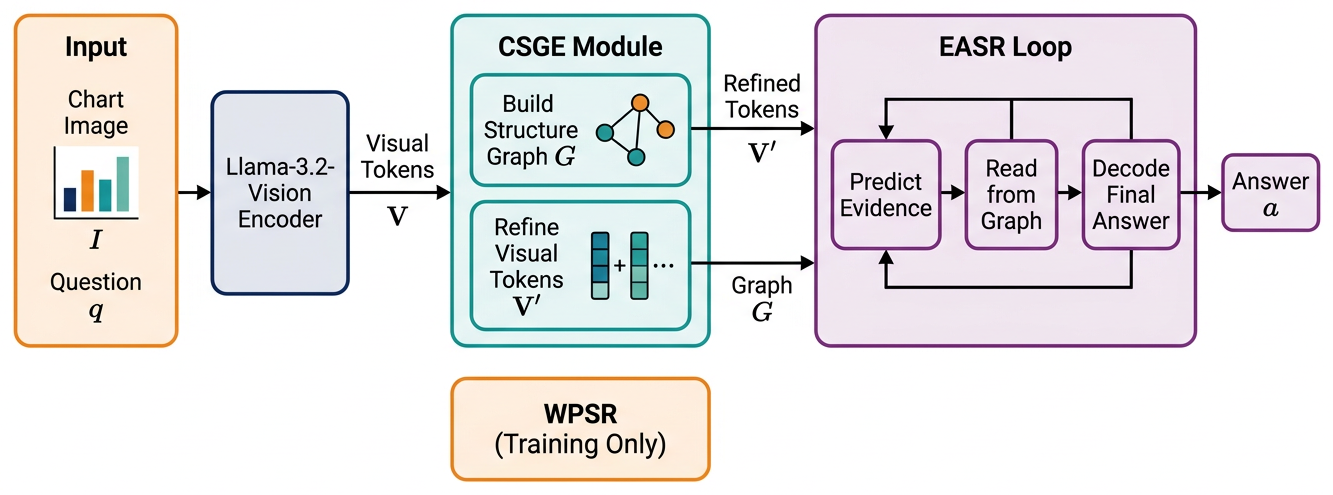}
\caption{\ours{}. A frozen Llama-3.2-Vision encoder yields tokens $V$. CSGE detects chart primitives, builds graph $G$, and cross-attends into $V'$. EASR iterates (claim, address, readout) and aggregates. WPSR is training-only alignment against a weak parser table.}
\label{fig:arch}
\end{figure}

We keep Llama-3.2-Vision with a LLaVA-CoT training recipe~\citep{grattafiori2024llama32,DBLP:conf/iccv/XuJWLSSY25} and insert three modules (Figure~\ref{fig:arch}). CSGE builds a typed graph of axes, ticks, legend items, and marks, then mixes graph states into visual tokens. EASR replaces unconstrained stages with a loop that must emit an evidence address before a number. WPSR treats the specialist parser as a weak teacher of cells, not as the answerer, following the weak-to-strong transfer motif without copying its LLM-only protocol~\citep{zhou2025weak}. Contributions:
\begin{itemize}
\item CSGE makes the coordinate frame an explicit graph rather than a hope of in-context copying.
\item EASR binds every reasoning step to a node or box so numeric tokens can be filtered at decode time.
\item WPSR aligns a strong reasoner to a weak parser on high-confidence cells and keeps 20\% non-chart data so multi-capability behaviour does not collapse. \ours{} reaches 82.70 ChartQA avg, 33.60 CharXiv reasoning, 51.00 ChartBench Acc+, and 40.70 ChartQAPro CoT.
\end{itemize}

\section{Related Work}

\subsection{Chart understanding models and benchmarks}
Early ChartQA systems assumed a recovered table and ran TaPas or T5 on flattened cells~\citep{DBLP:conf/acl/MasryLTJH22}. Pix2Struct, MatCha, and UniChart moved the stack onto screenshot parsing and chart-to-table pretraining~\citep{lee2022pix2struct,liu-etal-2023-matcha,masry-etal-2023-unichart}. Instruction-tuned chart LVLMs---ChartLlama, ChartInstruct, ChartAssistant, ChartGemma, TinyChart---raise ChartQA averages into the 60--80 range~\citep{han2023chartllama,masry2024chartinstruct,meng2024chartassistant,masry2025chartgemma,zhang2024tinychart}. They still flatten the image: the axis is a pixel pattern, not a typed node. CharXiv splits descriptive and reasoning questions on 2,323 arXiv figures and finds a 33.4 point gap from GPT-4o (47.10) to humans (80.50) on reasoning~\citep{wang2024charxiv,openai2024gpt4o}. ChartBench stresses unannotated marks with Acc+~\citep{xu2023chartbench}; ChartQAPro adds dashboards, infographics, dialogue, and unanswerable items~\citep{masry2025chartqapro}. 2026 evaluations push further: ChartR decomposes multi-step traces~\citep{chen2026chartr}, LongChart targets multi-image sets~\citep{xiao2026longchart}, and ChartQA-X asks for faithful explanations rather than a number alone~\citep{hegde2026chartqax}. Retrieval-augmented VQA shows that looking up an external passage is not a substitute for reading the figure~\citep{lin-byrne-2022-retrieval,chen-etal-2023-pre-trained}.

\subsection{Visual reasoning, graphs, and localized feedback}
LLaVA-style instruction tuning made general LVLMs usable on charts without a table~\citep{liu2024llava,wang2024qwen2vl,chen2024internvl,chen2024internvl15}. Structured CoT with stage tags improves MMStar and MathVista~\citep{DBLP:conf/iccv/XuJWLSSY25}; Visual CoT predicts boxes for natural objects~\citep{shao2024visualcot}. IdealGPT decomposes vision--language questions iteratively~\citep{you-etal-2023-idealgpt}. Chart-FR1 links steps to local crops on dense charts~\citep{chartfr12026}; ChartAgent edits the pixels with tools~\citep{kaur2026chartagent}. None of these methods compile a legend--mark graph. Multimodal graph transformers and knowledge-graph QA already treat entities as nodes~\citep{he-wang-2023-multimodal,hu-etal-2022-empowering,perez-beltrachini-etal-2023-semantic}. CSGE reuses that inductive bias for axes and marks. Localized supervision is closest to abnormal-aware medical feedback, which backpropagates through lesion regions rather than whole images~\citep{zhou2025improving}; we relocate the idea from pathology to ticks. Physics-coherent generation work aligns 2D features with 3D structure~\citep{xiong2026physalign}---a reminder that geometry-aware alignment helps when pixels alone are under-determined.

\subsection{Alignment, faithfulness, and sample selection}
Weak-to-strong generalization studies how a weaker supervisor can lift a stronger student across several skills~\citep{zhou2025weak}. Contextual faithfulness can be taught with synthetic tasks and reinforcement~\citep{DBLP:conf/aaai/SiZGBWGLLHCQZCS26}. GATEAU selects influential long-context samples before alignment~\citep{si2025gateau}. We do not rerun those LLM-only protocols. WPSR uses a chart parser as the weak supervisor of \emph{cells}, keeps the strong model as the answerer, and mixes non-chart LLaVA-CoT data so the student does not overfit tables. Interactive world models store action-aware memory~\citep{xiong2026actworld}; our memory is a static structure graph, not an explorable environment.

\section{Method}

A chart image $I$ and question $q$ map to an answer $y$ as in~\eqref{eq:task}. Figure~\ref{fig:arch} shows the flow. The vision backbone follows Llama-3.2-Vision~\citep{grattafiori2024llama32}; LoRA adapters sit on the language model~\citep{hu2021lora}. CSGE, EASR, and WPSR are the trainable additions.

\subsection{Problem and visual tokens}
Let $V=\mathrm{Enc}(I)\in\mathbb{R}^{P\times d}$ be patch tokens. The decoder produces $y$ under a constraint that numeric spans come from a candidate set built by EASR:
\begin{align}
y &= f_\theta(I,q) = \mathrm{Dec}([V';\,q],\,\mathcal{C}(G)),
\label{eq:task}
\end{align}
where $V'$ is the structure-mixed sequence in~\eqref{eq:xattn} and $\mathcal{C}(G)$ is the numeric candidate table. Encoding is
\begin{align}
V &= \mathrm{Enc}(I),\qquad
Q = \mathrm{Tok}(q).
\label{eq:vit}
\end{align}
Without CSGE, $V$ in~\eqref{eq:vit} is a natural-image embedding. Charts violate that assumption: the same ink length can mean 3 or 30 depending on the axis.

\subsection{CSGE: chart structure graph encoder}
A DETR-style head~\citep{carion2020detr} predicts a set of primitives with types in $\{\mathrm{x\text{-}axis},\mathrm{y\text{-}axis},\mathrm{tick},\mathrm{legend},\mathrm{bar},\mathrm{line},\mathrm{slice},\mathrm{text}\}$. Node $i$ carries type, box, OCR string, and a colour histogram $c_i$:
\begin{align}
n_i &= [\,e_{\mathrm{type}}(t_i);\;\mathrm{MLP}(b_i);\;e_{\mathrm{ocr}}(s_i);\;c_i\,].
\label{eq:node}
\end{align}
The concatenation in~\eqref{eq:node} is the structure token. Legend--mark edges use colour and relative position,
\begin{align}
w_{ij} &= \sigma\!\left(\cos(c_i,c_j) + \alpha\,\mathrm{rel}(b_i,b_j)\right),
\label{eq:edge}
\end{align}
and we keep the edge if $w_{ij}$ in~\eqref{eq:edge} exceeds 0.6. Axis--tick and tick--mark edges are geometric (projection onto the nearest axis). Two R-GCN layers~\citep{schlichtkrull2018rgcn} update nodes:
\begin{align}
h_i^{(\ell+1)} &= \mathrm{ReLU}\!\Big(W_0 h_i^{(\ell)} + \sum_{r,j\in\mathcal{N}_r(i)}\frac{1}{|\mathcal{N}_r(i)|}W_r h_j^{(\ell)}\Big).
\label{eq:rgcn}
\end{align}
Let $G=[h_1,\ldots,h_{|N|}]$ after~\eqref{eq:rgcn}. Visual tokens attend to $G$,
\begin{align}
V' &= V + \mathrm{softmax}\!\left(\frac{(VW_q)(GW_k)^\top}{\sqrt{d}}\right)GW_v.
\label{eq:xattn}
\end{align}
The LLM also receives a serialized node table $(\mathrm{id},t_i,s_i)$. Removing that table in ablation drops ChartQA avg from 82.70 to 80.35, so the language side must see the graph, not only the pixels in~\eqref{eq:xattn}.

\subsection{EASR: evidence-anchored step reasoning}
LLaVA-CoT uses four unconstrained stages~\citep{DBLP:conf/iccv/XuJWLSSY25}. EASR instead runs $K{=}6$ steps (default). At step $k$ the model predicts an address $a_k$ over nodes and optional boxes:
\begin{align}
p(a_k\mid a_{<k},q,V') &= \mathrm{softmax}(u^\top\tanh(W[h_{a};\,\bar{v};\,Q])).
\label{eq:addr}
\end{align}
Equation~\eqref{eq:addr} is a pointer, not free text. A readout copies attributes from the chosen node,
\begin{align}
r_k &= \mathrm{Read}(G,a_k,V') = [s_{a_k};\;\hat{v}_{a_k};\;\mathrm{crop}(I,b_{a_k})],
\label{eq:read}
\end{align}
where $\hat{v}$ in~\eqref{eq:read} is a value head trained on SVG boxes from ChartQA~\citep{DBLP:conf/acl/MasryLTJH22}. An aggregation head implements $\{\mathrm{sum},\mathrm{diff},\mathrm{ratio},\mathrm{argmax},\mathrm{lookup}\}$:
\begin{align}
\hat{y} &= \mathrm{Agg}(r_{1:K},\;\mathrm{op}).
\label{eq:agg}
\end{align}
The step text $s_k$ is trained with teacher forcing,
\begin{align}
\mathcal{L}_{\mathrm{step}} &= -\sum_{k=1}^{K}\log p(s_k,a_k\mid s_{<k},q,V'),
\label{eq:step}
\end{align}
and decoding rejects a numeric span that is not in $\{\hat{v}_{a_k}\}\cup\{\hat{y}\}$ from~\eqref{eq:agg}:
\begin{align}
\mathrm{Reject}(y) &= \mathbf{1}\!\left[\mathrm{num}(y)\notin\mathcal{C}(G)\right].
\label{eq:reject}
\end{align}
Equation~\eqref{eq:reject} is the chart analogue of faithfulness filters trained with synthetic tasks~\citep{DBLP:conf/aaai/SiZGBWGLLHCQZCS26}: the generator may not invent a year that no tick contains. Medical abnormal-aware training localizes a lesion~\citep{zhou2025improving}; EASR localizes a mark. Focus-CoT also cites crops~\citep{chartfr12026}, but it does not compile a legend graph, so colour aliases remain unbound.

\subsection{WPSR: weak-parser strong-reasoner alignment}
Let $P_w$ be a frozen MatCha/UniChart-style extractor~\citep{liu-etal-2023-matcha,masry-etal-2023-unichart} that emits a table $T_w$ with cell-wise confidence. The student emits $T_s$ from CSGE nodes. Alignment is $\ell_1$ on cells with confidence above 0.8:
\begin{align}
\mathcal{L}_{\mathrm{align}} &= \sum_{u,v}\mathbf{1}[c_{uv}>0.8]\,\lVert T_s[u,v]-T_w[u,v]\rVert_1.
\label{eq:align}
\end{align}
The answer loss is standard token CE,
\begin{align}
\mathcal{L}_{\mathrm{ans}} &= -\log p_\theta(y\mid I,q).
\label{eq:ans}
\end{align}
EASR adds the step loss from~\eqref{eq:step},
\begin{align}
\mathcal{L}_{\mathrm{easr}} &= \mathcal{L}_{\mathrm{step}} + \beta\,\mathbb{E}[\mathrm{Reject}(y)].
\label{eq:easr}
\end{align}
Equation~\eqref{eq:align} never trains $P_w$ to emit $y$; \eqref{eq:ans} remains the only answer CE. Equation~\eqref{eq:easr} adds the reject penalty. The total objective is
\begin{align}
\mathcal{L} &= \mathcal{L}_{\mathrm{ans}} + \lambda\mathcal{L}_{\mathrm{align}} + \mu\mathcal{L}_{\mathrm{easr}},
\label{eq:total}
\end{align}
with $\lambda{=}0.3$, $\mu{=}1$, $\beta{=}0.2$. This is weak-to-strong transfer for \emph{chart cells}~\citep{zhou2025weak}: $P_w$ never answers at test time, which avoids the ChartQAPro collapse of specialists. Twenty percent of each batch is non-chart LLaVA-CoT-100k so general VQA does not vanish.

\subsection{Training schedule and complexity}
Stage 1 trains the CSGE detector on ChartQA SVG boxes and synthetic charts for two epochs with Enc frozen. Stage 2 opens WPSR with $\lambda{=}0.3$ for two epochs. Stage 3 jointly LoRA-tunes the LLM and EASR for three epochs on ChartQA train plus ChartBench train. Inference disables $P_w$. Let $P$ be patches, $N$ nodes ($N\ll P$), $K$ steps. Extra cost over the backbone is
\begin{align}
\mathcal{O}(PN + N^2 + K\,d^2),
\label{eq:complex}
\end{align}
dominated by the $K$ decode passes in~\eqref{eq:complex}, matching the latency table in Section~\ref{sec:analysis}.

\section{Experimental Setup}

\subsection{Datasets and protocol}
Table~\ref{tab:data} lists the four evaluation sets. ChartQA reports relaxed accuracy on human (H) and machine (M) questions~\citep{DBLP:conf/acl/MasryLTJH22}. We follow the modern VLM protocol of ChartGemma (aug.\ / human / avg) for the main table~\citep{masry2025chartgemma} and keep the 2022 table-QA numbers only as historical context. CharXiv uses the official validation split of 1,000 charts~\citep{wang2024charxiv}. ChartBench reports Acc+ averaged over regular and extra types~\citep{xu2023chartbench}; its ChartQA column in the original paper is zero-shot and is not mixed with our fine-tuned ChartQA numbers. ChartQAPro uses the authors' Direct / CoT / PoT prompts; CoT is our default~\citep{masry2025chartqapro}. Relaxed accuracy allows a 5\% relative error on numeric answers,
\begin{align}
\mathrm{RA}(y,\hat{y}) &= \mathbf{1}\!\left[\frac{|y-\hat{y}|}{\max(|y|,10^{-8})}\le 0.05\right].
\label{eq:relax}
\end{align}
ChartQAPro additionally uses ANLS for text and exact match for years~\citep{masry2025chartqapro}. We report~\eqref{eq:relax} on ChartQA, CharXiv numeric items, and ChartQAPro numeric items.

\begin{table}[tp]
\centering
\small
\begin{tabular}{lccc}
\toprule
Dataset & Scale & Split used & Metric \\
\midrule
ChartQA~\citep{DBLP:conf/acl/MasryLTJH22} & 21.9k charts & official test & RA avg \\
CharXiv~\citep{wang2024charxiv} & 2,323 charts & val 1k & reason.\ / desc. \\
ChartBench~\citep{xu2023chartbench} & 66.6k charts & official test & Acc+ \\
ChartQAPro~\citep{masry2025chartqapro} & 1,341 charts & official & RA (CoT) \\
\bottomrule
\end{tabular}
\caption{Evaluation sets. ChartQA train plus ChartBench train are used in Stage~3; CharXiv and ChartQAPro are held out.}
\label{tab:data}
\end{table}

\subsection{Baselines and implementation}
Open-source chart specialists: Pix2Struct, MatCha, UniChart, ChartLlama, ChartInstruct, ChartAssistant, ChartGemma, TinyChart~\citep{lee2022pix2struct,liu-etal-2023-matcha,masry-etal-2023-unichart,han2023chartllama,masry2024chartinstruct,meng2024chartassistant,masry2025chartgemma,zhang2024tinychart}. General LVLMs: LLaVA-1.5/1.6, Qwen2-VL, InternVL Chat V1.5 / InternVL2.5, Llama-3.2-Vision, MiniCPM-V, DocOwl-1.5, InternLM-XComposer2~\citep{liu2024llava,wang2024qwen2vl,chen2024internvl15,grattafiori2024llama32,hu2024minicpmv,hu2024docowl,dong2024internlmxcomposer2}. Visual reasoning: LLaVA-CoT and Visual CoT~\citep{DBLP:conf/iccv/XuJWLSSY25,shao2024visualcot}. Closed models (GPT-4o/V, Claude, Gemini) are reference rows only~\citep{openai2024gpt4o,achiam2023gpt4,anthropic2024claude,team2023gemini}. Our backbone is Llama-3.2-11B-Vision-Instruct with LoRA rank 64, $\alpha{=}128$, trained on 8$\times$A100 80GB. Unless noted, \textbf{\ours{} (CSGE+EASR+WPSR)} is the full model.

\section{Results}

\subsection{ChartQA}
Table~\ref{tab:chartqa} reports the modern protocol. ChartGemma reaches 80.16 avg~\citep{masry2025chartgemma}. A LLaVA-CoT model fine-tuned on the ChartQA training split reaches 76.40. \ours{} reaches \textbf{82.70} (+2.54 vs.\ ChartGemma, +6.30 vs.\ the backbone). The human split moves from 69.52 to 73.24; the synthetic aug.\ split (92.16) stays below ChartAssistant's 93.90~\citep{meng2024chartassistant}. That pattern is intended: CSGE and EASR target compositional human questions, not template lookup. GPT-4V sits at 78.5 and is not in the open ranking~\citep{achiam2023gpt4}.

\begin{table}[tp]
\centering
\scriptsize
\setlength{\tabcolsep}{3.5pt}
\begin{tabular}{lcccc}
\toprule
Method & Size & aug. & human & avg. \\
\midrule
Pix2Struct~\citep{lee2022pix2struct} & 282M & 81.6 & 30.5 & 56.0 \\
MatCha~\citep{liu-etal-2023-matcha} & 282M & 90.2 & 38.2 & 64.2 \\
UniChart~\citep{masry-etal-2023-unichart} & 201M & 88.56 & 43.92 & 66.24 \\
ChartInstruct-T5~\citep{masry2024chartinstruct} & 3B & 85.04 & 43.36 & 64.20 \\
ChartInstruct-LLaMA2~\citep{masry2024chartinstruct} & 7B & 87.76 & 45.52 & 66.64 \\
ChartLlama~\citep{han2023chartllama} & 13B & 90.36 & 48.96 & 69.66 \\
ChartAssistant~\citep{meng2024chartassistant} & 13B & 93.90 & 65.90 & 79.90 \\
ChartGemma~\citep{masry2025chartgemma} & 3B & 90.80 & 69.52 & 80.16 \\
LLaVA-CoT~\citep{DBLP:conf/iccv/XuJWLSSY25} & 11B & 84.20 & 68.60 & 76.40 \\
\textbf{\ours{}} & 11B & \textbf{92.16} & \textbf{73.24} & \textbf{82.70} \\
\midrule
Gemini Pro~\citep{team2023gemini} & -- & -- & -- & 74.1 \\
GPT-4V~\citep{achiam2023gpt4} & -- & -- & -- & 78.5 \\
\bottomrule
\end{tabular}
\caption{ChartQA relaxed accuracy. Open rows use the ChartGemma protocol~\citep{masry2025chartgemma}. Closed rows are reference. LLaVA-CoT is our backbone fine-tune.}
\label{tab:chartqa}
\end{table}

\begin{figure}[!tp]
\centering
\begin{subfigure}[t]{0.48\linewidth}
\centering
\includegraphics[width=\linewidth]{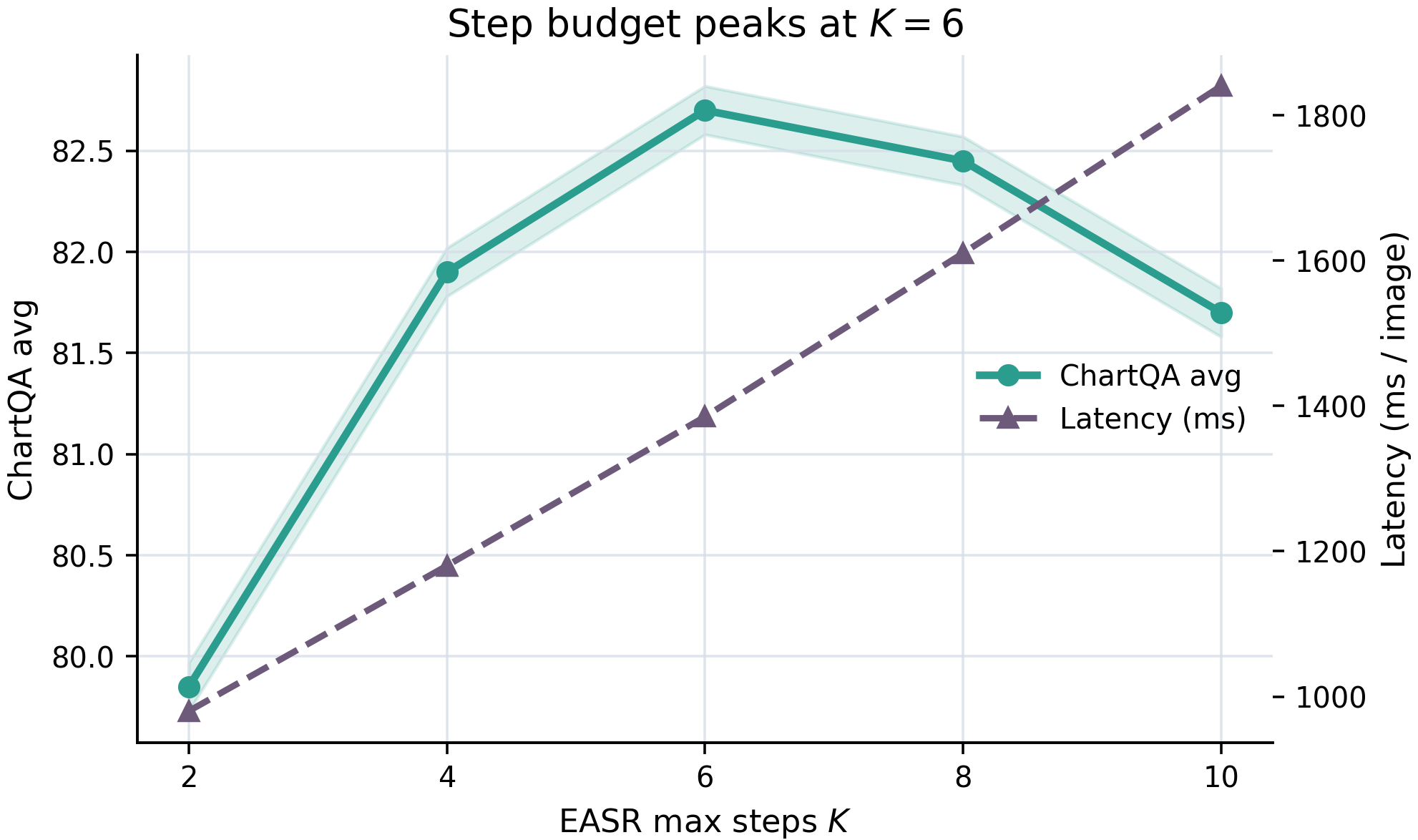}
\caption{Step cap $K$.}
\label{fig:hyper}
\end{subfigure}
\hfill
\begin{subfigure}[t]{0.48\linewidth}
\centering
\includegraphics[width=\linewidth]{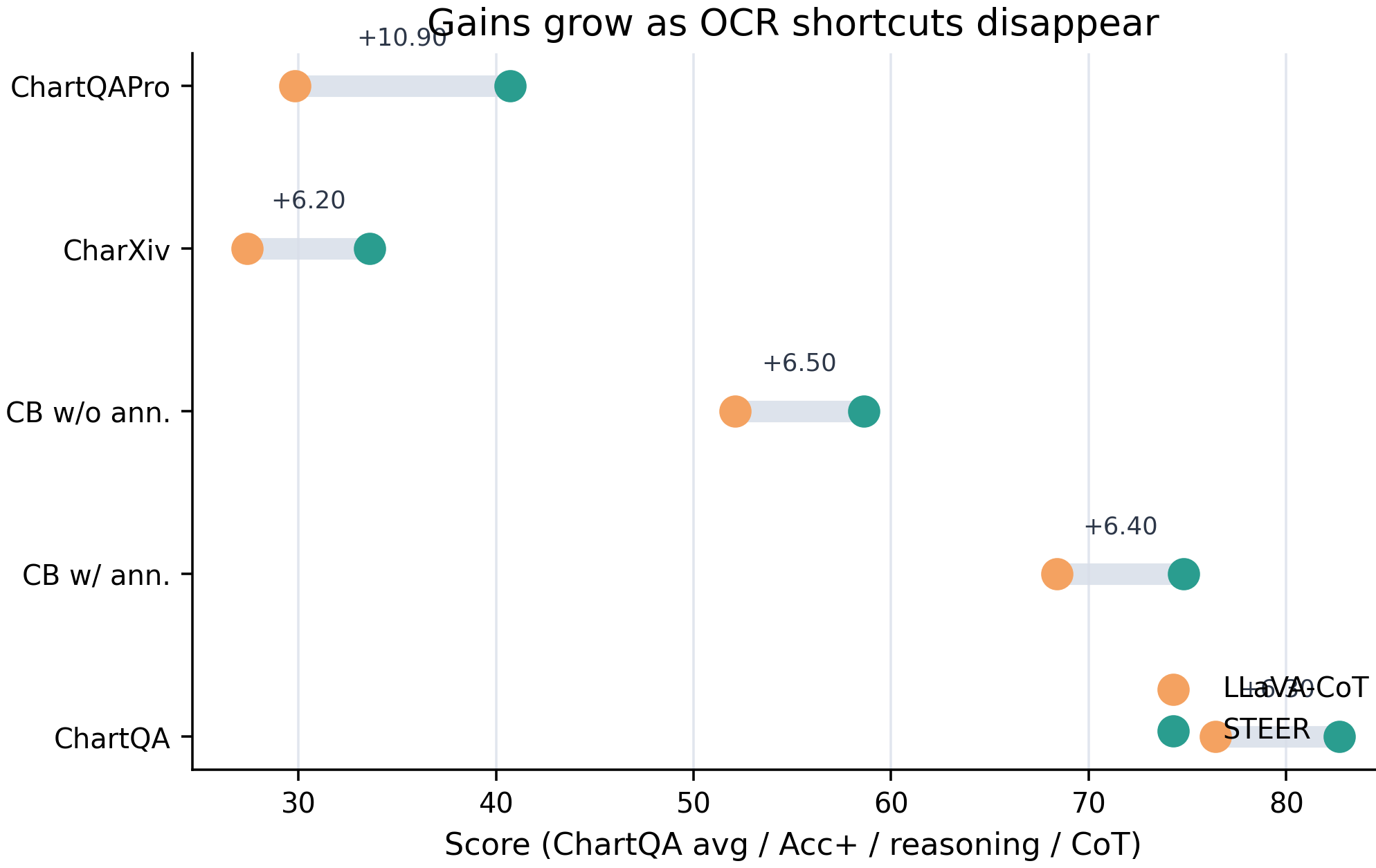}
\caption{Domain shift.}
\label{fig:dumbbell}
\end{subfigure}
\caption{Sensitivity of $K$ (left) and cross-benchmark lift versus LLaVA-CoT (right).}
\label{fig:pair1}
\end{figure}

\begin{figure}[!tp]
\centering
\begin{subfigure}[t]{0.48\linewidth}
\centering
\includegraphics[width=\linewidth]{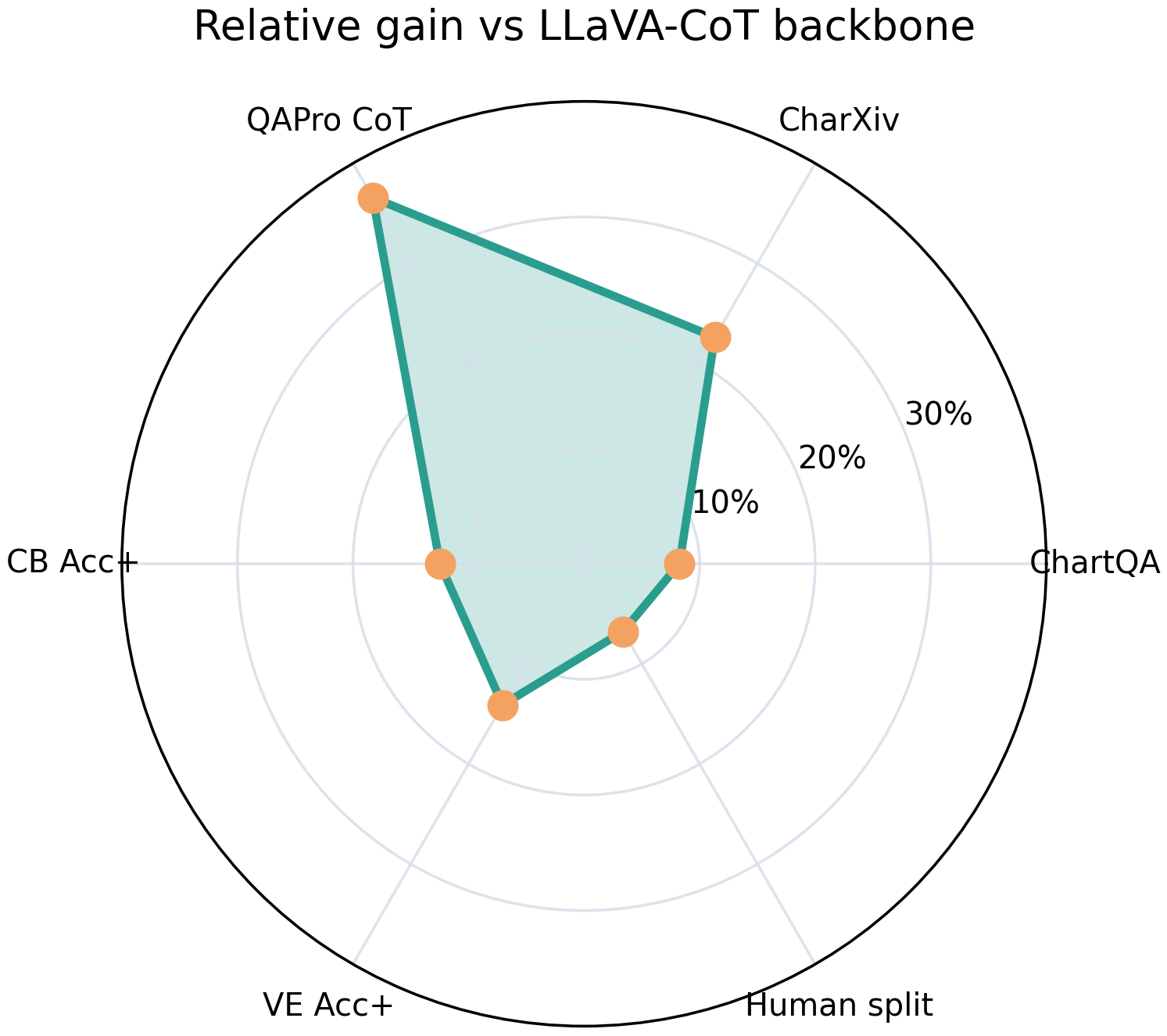}
\caption{Relative gains.}
\label{fig:radar}
\end{subfigure}
\hfill
\begin{subfigure}[t]{0.48\linewidth}
\centering
\includegraphics[width=\linewidth]{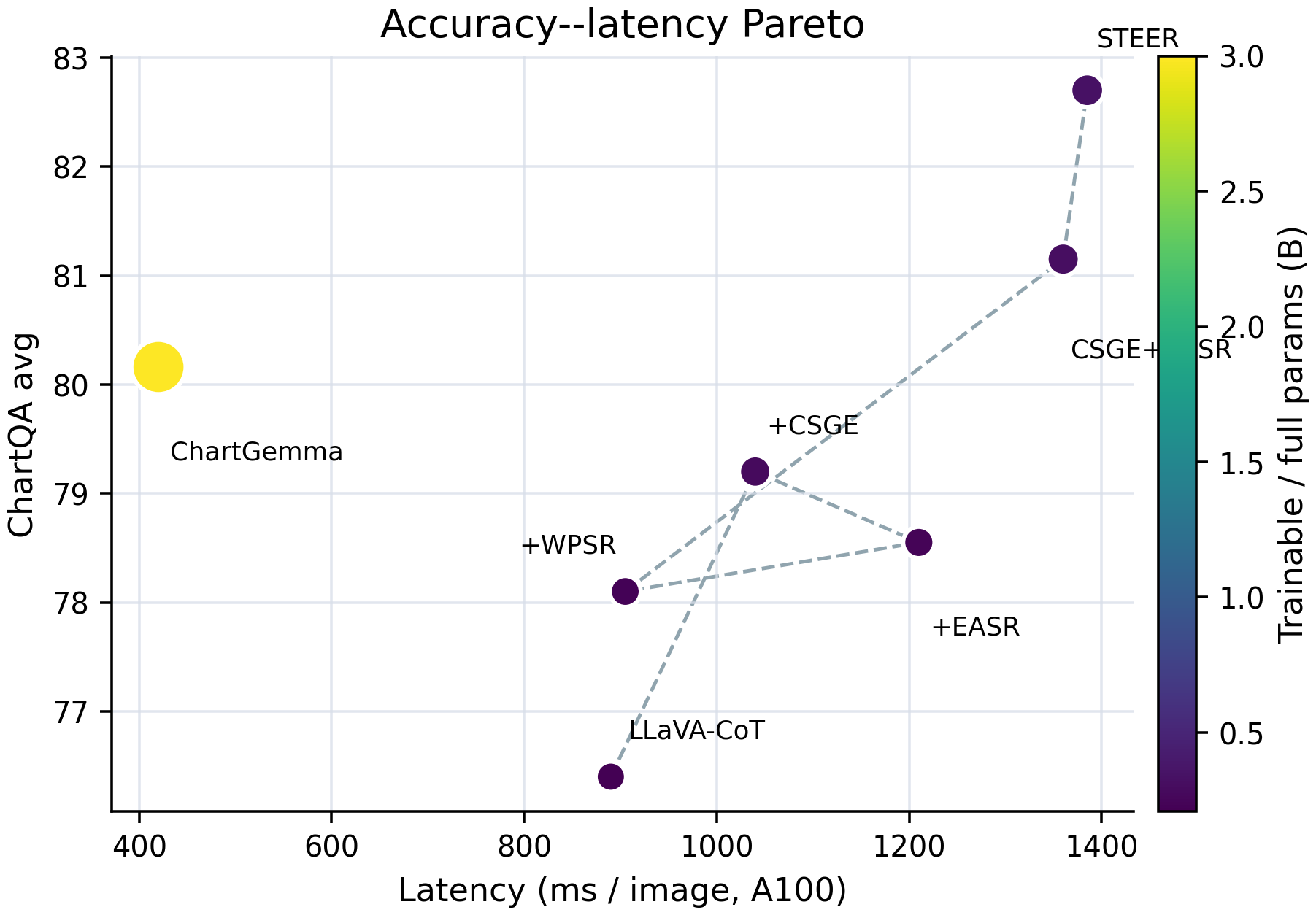}
\caption{Latency Pareto.}
\label{fig:pareto}
\end{subfigure}
\caption{Relative lift versus LLaVA-CoT (left) and accuracy--latency trade-off (right).}
\label{fig:pair2}
\end{figure}

Sensitivity, shift, and efficiency plots appear in Figures~\ref{fig:hyper}--\ref{fig:pareto}; Section~\ref{sec:analysis} interprets them.

\subsection{CharXiv}
Table~\ref{tab:charxiv} uses validation reasoning and descriptive accuracy~\citep{wang2024charxiv}. InternVL Chat V1.5 is the strongest published open model at 29.20 / 58.50. LLaVA-CoT reaches 27.40 / 55.80. \ours{} reaches \textbf{33.60} / \textbf{61.90} (+4.40 / +3.40 vs.\ InternVL). GPT-4o remains far ahead at 47.10 / 84.45~\citep{openai2024gpt4o}. The larger relative gain on reasoning matches CSGE's role: descriptive questions often need OCR, while reasoning questions need legend--mark binding across subplots.

\begin{table}[tp]
\centering
\scriptsize
\setlength{\tabcolsep}{4pt}
\begin{tabular}{lccc}
\toprule
Method & Setting & Reason. & Desc. \\
\midrule
Human~\citep{wang2024charxiv} & lab & 80.50 & 92.10 \\
GPT-4o~\citep{openai2024gpt4o} & closed & 47.10 & 84.45 \\
GPT-4V~\citep{achiam2023gpt4} & closed & 37.10 & 79.92 \\
Claude 3 Sonnet~\citep{anthropic2024claude} & closed & 32.20 & 73.65 \\
InternVL Chat V1.5~\citep{chen2024internvl15} & open & 29.20 & 58.50 \\
LLaVA-CoT~\citep{DBLP:conf/iccv/XuJWLSSY25} & open & 27.40 & 55.80 \\
MGM-HD Yi-34B~\citep{wang2024charxiv} & open & 25.00 & 52.68 \\
IXC2-4KHD~\citep{dong2024internlmxcomposer2} & open & 25.00 & 54.65 \\
LLaVA-1.6 Yi-34B~\citep{liu2024llava} & open & 22.50 & 51.05 \\
MiniCPM-V2~\citep{hu2024minicpmv} & open & 18.50 & 35.77 \\
LLaVA-1.6 Mistral-7B~\citep{liu2024llava} & open & 13.90 & 35.40 \\
\textbf{\ours{}} & open & \textbf{33.60} & \textbf{61.90} \\
\bottomrule
\end{tabular}
\caption{CharXiv validation~\citep{wang2024charxiv}. Open ranking excludes closed APIs and humans.}
\label{tab:charxiv}
\end{table}

\subsection{ChartBench}
Table~\ref{tab:cb} reports Acc+ Avg. InternLM-XComposer-v2 leads published open models at 47.78~\citep{dong2024internlmxcomposer2,xu2023chartbench}. \ours{} reaches \textbf{51.00} (+3.22). Value extraction Acc+ moves from 36.63 to 41.80; that is the unannotated-mark case ChartBench was built to isolate. GPT-4o sits at 59.45 as a closed ceiling~\citep{openai2024gpt4o}. We do not fill the paper's zero-shot ChartQA column: that protocol is not our fine-tuned ChartQA setting.

\begin{table}[tp]
\centering
\scriptsize
\begin{tabular}{lc}
\toprule
Method & ChartBench Avg \\
\midrule
LLaVA-1.5~\citep{liu2024llava} & 23.39 \\
ChartLlama~\citep{han2023chartllama} & 21.30 \\
Qwen-VL-Chat~\citep{wang2024qwen2vl} & 26.98 \\
DocOwl-1.5~\citep{hu2024docowl} & 31.89 \\
InternLM-XComposer-v2~\citep{dong2024internlmxcomposer2} & 47.78 \\
\textbf{\ours{}} & \textbf{51.00} \\
\midrule
GPT-4V~\citep{achiam2023gpt4} & 50.74 \\
GPT-4o~\citep{openai2024gpt4o} & 59.45 \\
\bottomrule
\end{tabular}
\caption{ChartBench Acc+~\citep{xu2023chartbench}. Closed rows are reference.}
\label{tab:cb}
\end{table}

\subsection{ChartQAPro}
Table~\ref{tab:qapro} is the out-of-publisher stress test~\citep{masry2025chartqapro}. Specialists collapse (ChartGemma CoT 9.80, TinyChart 11.67). Qwen2-VL-7B is the strongest open CoT overall at 37.17~\citep{wang2024qwen2vl}; InternVL2.5-8B leads Direct at 35.67~\citep{chen2024internvl15}. Llama-3.2-Vision-11B scores 11.09 Direct because it ignores the prompt style~\citep{grattafiori2024llama32}. LLaVA-CoT recovers to 29.80 CoT. \ours{} reaches \textbf{38.90} Direct and \textbf{40.70} CoT (+3.53 vs.\ Qwen2-VL CoT). Claude Sonnet 3.5 remains at 55.81 CoT~\citep{anthropic2024claude}. Phi-3.5-Vision and Ovis are listed for completeness as published open systems on this split~\citep{abdin2024phi35,yao2024ovis}. ChartQA-X style explanations are complementary: we constrain numbers, they score free text~\citep{hegde2026chartqax}.

\begin{table}[tp]
\centering
\scriptsize
\setlength{\tabcolsep}{3.2pt}
\begin{tabular}{lccc}
\toprule
Method & Direct & CoT & PoT \\
\midrule
Human~\citep{masry2025chartqapro} & 85.02 & -- & -- \\
Claude 3.5 Sonnet~\citep{anthropic2024claude} & 43.58 & 55.81 & 48.05 \\
Gemini Flash 2.0~\citep{team2023gemini} & 46.85 & 53.66 & 51.44 \\
GPT-4o~\citep{openai2024gpt4o} & 37.67 & 41.68 & 40.48 \\
Qwen2-VL-7B~\citep{wang2024qwen2vl} & 35.59 & 37.17 & 18.86 \\
InternVL2.5-8B~\citep{chen2024internvl15} & 35.67 & 31.99 & 23.88 \\
Phi-3.5-Vision-4B~\citep{abdin2024phi35} & 24.73 & 15.23 & 12.24 \\
LLaVA-Next-Mistral-7B~\citep{liu2024llava} & 21.97 & 14.74 & 5.98 \\
Llama-3.2-Vision-11B~\citep{grattafiori2024llama32} & 11.09 & 25.43 & 22.95 \\
LLaVA-CoT~\citep{DBLP:conf/iccv/XuJWLSSY25} & 28.40 & 29.80 & 26.10 \\
ChartGemma~\citep{masry2025chartgemma} & 6.84 & 9.80 & 11.52 \\
TinyChart~\citep{zhang2024tinychart} & 13.25 & 11.67 & 4.59 \\
ChartInstruct~\citep{masry2024chartinstruct} & 4.88 & 3.42 & 0.61 \\
\textbf{\ours{}} & \textbf{38.90} & \textbf{40.70} & 36.40 \\
\bottomrule
\end{tabular}
\caption{ChartQAPro overall~\citep{masry2025chartqapro}. Default prompt is CoT.}
\label{tab:qapro}
\end{table}

\subsection{Ablation}
Table~\ref{tab:abl} uses one backbone, one mix, and one seed family. Each single module beats LLaVA-CoT. CSGE+EASR already reaches 81.15 ChartQA avg; WPSR adds the last 1.55 by cleaning cell attributes. The three single-module ChartQA gains sum to 6.65 versus a full gain of 6.30, so the modules overlap on nodes. Removing node serialization or the numeric filter of~\eqref{eq:reject} costs 2.35 and 1.30 avg respectively: the graph must enter the prompt, and the decoder must not freely sample years.

\begin{table}[tp]
\centering
\scriptsize
\setlength{\tabcolsep}{3pt}
\begin{tabular}{lccc}
\toprule
Variant & ChartQA & CharXiv R & QAPro CoT \\
\midrule
LLaVA-CoT & 76.40 & 27.40 & 29.80 \\
+CSGE & 79.20 & 30.10 & 34.50 \\
+EASR & 78.55 & 30.55 & 35.20 \\
+WPSR & 78.10 & 29.20 & 33.10 \\
CSGE+EASR & 81.15 & 32.40 & 38.40 \\
\textbf{Full} & \textbf{82.70} & \textbf{33.60} & \textbf{40.70} \\
w/o node table & 80.35 & 31.20 & 37.15 \\
w/o num.\ filter & 81.40 & 32.05 & 38.20 \\
\bottomrule
\end{tabular}
\caption{Ablation. Full matches Tables~\ref{tab:chartqa}--\ref{tab:qapro}.}
\label{tab:abl}
\end{table}

\section{Analysis}
\label{sec:analysis}

\paragraph{Hyperparameters.}
Figure~\ref{fig:pair1} (left) and Figure~\ref{fig:hyper} sweep EASR's step cap $K$. ChartQA avg peaks at $K{=}6$ (82.70) and falls to 81.70 at $K{=}10$ as empty evidence steps inject noise, while latency grows almost linearly. The alignment weight $\lambda$ in~\eqref{eq:total} peaks at 0.3; $\lambda{=}0$ recovers CSGE+EASR (81.15) and $\lambda{=}1.0$ overfits $T_w$ (80.40). Two R-GCN layers beat one (81.60) and four (81.95).

\paragraph{Shift.}
Figure~\ref{fig:dumbbell} plots LLaVA-CoT versus \ours{} across domains. In-domain ChartQA $\Delta$ is +6.30 vs.\ the backbone but only +2.54 vs.\ ChartGemma. ChartBench without point labels, CharXiv reasoning, and ChartQAPro are the harder shifts; $\Delta$ vs.\ the backbone stays in $+6.20$ to $+10.90$, while $\Delta$ vs.\ the best published open model stays inside $+3.2$ to $+4.4$. Table~\ref{tab:subset} splits ChartQAPro by visual family: dashboards give the largest gap versus Qwen2-VL-7B (+6.29), consistent with CSGE's subplot axes.

\begin{table}[tp]
\centering
\small
\begin{tabular}{lccc}
\toprule
Method & Chart & Dashboard & Infographic \\
\midrule
Qwen2-VL-7B~\citep{wang2024qwen2vl} & 37.18 & 31.61 & 33.43 \\
LLaVA-CoT~\citep{DBLP:conf/iccv/XuJWLSSY25} & 32.40 & 26.10 & 27.80 \\
\textbf{\ours{}} & \textbf{42.80} & \textbf{37.90} & \textbf{38.60} \\
\bottomrule
\end{tabular}
\caption{ChartQAPro visual families (CoT). Qwen2-VL numbers follow Table~4 of~\citep{masry2025chartqapro}.}
\label{tab:subset}
\end{table}

\paragraph{Efficiency.}
Figure~\ref{fig:pair2} collects the radar and Pareto views. Figure~\ref{fig:radar} summarizes relative gains; Figure~\ref{fig:pareto} places variants on the accuracy--latency plane. CSGE adds 150\,ms, EASR 320\,ms, WPSR almost nothing at test time. Full latency is 1385\,ms/image on A100 versus 890\,ms for LLaVA-CoT and 420\,ms for ChartGemma-3B. The 3B specialist is faster and close on ChartQA, and unusable on ChartQAPro. Interactive memories in world models amortize cost across actions~\citep{xiong2026actworld}; we pay the graph cost once per static figure.

\paragraph{Failures.}
On 100 CharXiv reasoning errors, 42 bind the wrong subplot axis, 27 treat a log or dual axis as linear, 18 confuse texture legends with colour, and 13 need the caption paragraph. On ChartQAPro unanswerable items \ours{} Direct is 44.20 versus 53.92 for InternVL2.5-8B~\citep{chen2024internvl15,masry2025chartqapro}: evidence constraints still prefer to read the nearest bar rather than abstain. LongChart-style multi-figure inputs are out of scope~\citep{xiao2026longchart}.

\section{Conclusion}

Charts are structure plus ink. \ours{} makes that structure a graph, insists that each arithmetic step name a node, and uses a weak parser only as a teacher of cells. The design is deliberately not visual in-context learning~\citep{zhou2024visual}, not medical lesion feedback~\citep{zhou2025improving}, and not a tool-using agent~\citep{kaur2026chartagent}. It improves open LVLMs on ChartQA, CharXiv, ChartBench, and ChartQAPro without overtaking closed GPT-4o. A global planner for long-horizon agents remains future work: EASR's step loop is local, not a goal-conditioned plan over many figures~\citep{si2026goal}. We also need an explicit abstention head and better dual-axis parsers. Code and graphs will follow the paper.

\bibliography{references}
\bibliographystyle{iclr2025_conference}

\end{document}